\documentclass[letterpaper, 10 pt, conference]{ieeeconf}  

\IEEEoverridecommandlockouts                              
\usepackage{graphicx} 
\usepackage{times} 
\usepackage{amsmath} 
\usepackage{amssymb}  
\usepackage{amsfonts}
\usepackage{color}
\usepackage{cite}
\usepackage[ruled,linesnumbered,vlined]{algorithm2e}
\usepackage{algpseudocode}

\title{\LARGE \bf
Multi-Vehicle Interaction Scenarios Generation with Interpretable Traffic Primitives and Gaussian Process Regression
}

\author{Weiyang Zhang, Wenshuo Wang, {\it Member, IEEE}, and Ding Zhao
\thanks{Corresponding Authors: Wenshuo Wang and Ding Zhao.}
\thanks{W. Zhang, W. Wang, and D. Zhao are with the Safe AI Lab, Carnegie Mellon University (CMU), Pittsburgh, PA 15213, USA. e-mail: {\tt\small zhangwy@umich.edu, wwsbit@gmail.com, dingzhao@cmu.edu}}
}

\begin{document}

\maketitle
\thispagestyle{empty}
\pagestyle{empty}

\begin{abstract}

Generating multi-vehicle interaction scenarios can benefit motion planning and decision making of autonomous vehicles when on-road data is insufficient. This paper presents an efficient approach to generate varied multi-vehicle interaction scenarios that can both adapt to different road geometries and inherit the key interaction patterns in  real-world driving. Towards this end, the available multi-vehicle interaction scenarios are temporally segmented into several interpretable fundamental building blocks, called {\it\textbf{traffic primitives}}, via the Bayesian nonparametric learning. Then, the changepoints of traffic primitives are transformed into the desired road to generate collision-free interaction trajectories through a sampling-based path planning algorithm. The Gaussian process regression is finally introduced to control the variance and smoothness of the generated multi-vehicle interaction trajectories. Experiments with simulation results of three typical multi-vehicle trajectories at different road conditions are carried out.  The experimental results demonstrate that our proposed method can generate a bunch of human-like multi-vehicle interaction trajectories that can fit different road conditions remaining the key interaction patterns of agents in the provided scenarios, which is import to the development of autonomous vehicles. 

\end{abstract}

\section{Introduction}

Autonomous vehicles will help create a safer, cleaner, and more mobile society and many researchers are contributing to develop fully autonomous driving systems \cite{mervis2017not}. Significant autonomous driving competition events such as DARPA challenge and Hyundai Autonomous Challenge have been held \cite{broggi2013extensive}. Industrial research has also accelerated this pace by developing several platforms such as Waymo, Toyota, and Baidu Apollo Driving Platforms \cite{luettel2012autonomous, rosenzweig2015review, bimbraw2015autonomous}. However, it is still far from achieving the goal \cite{levinson2011towards}. 

One of the biggest challenges lies in the proper interaction with human drivers in complex driving scenarios \cite{wang2018clustering}. Currently, the widely-used motion planning algorithms in autonomous vehicles mainly aim at generating safe, optimal, and computational feasible trajectories \cite{paden2016survey}, which can be classified as graph-based planners (e.g., A$^\ast$, state lattices), sampling-based planners (e.g., probabilistic roadmap, RRT$^\ast$), geometric-based planners (e.g., visibility graph), and optimization-based planners (e.g., model predictive planning and constrained optimization) \cite{gonzalez2015review, paden2016survey, katrakazas2015real}. 
\begin{figure}[t]
\centering
\includegraphics[width=\linewidth]{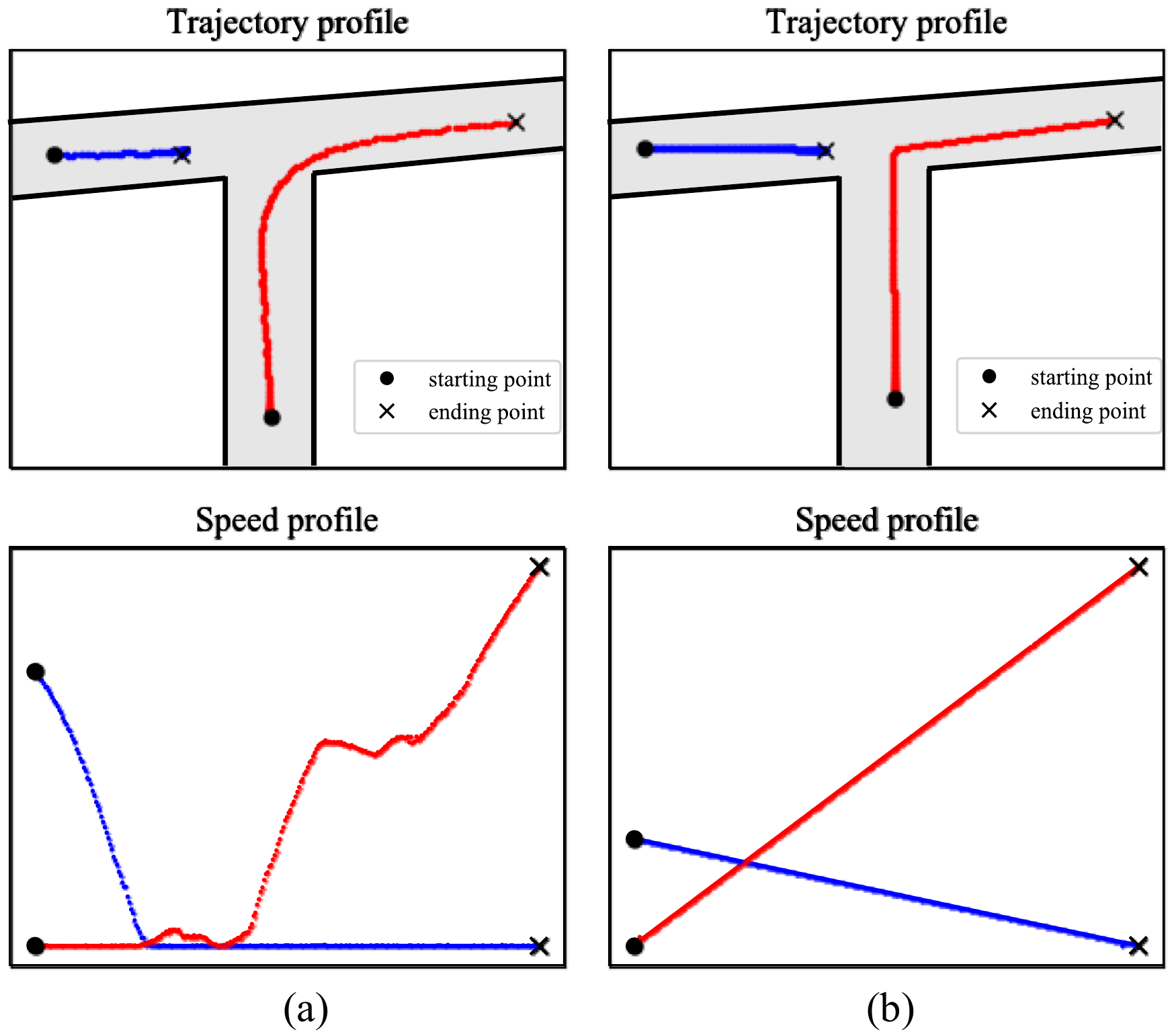}
\caption{A comparison between (a) naturalistic driving scenarios and (b) the scenarios generated by a simple path-planning algorithm.}
\label{fig:compare}
\end{figure}
However, a fully autonomous vehicle is expected to not only drive safely but also make human-like motions such that seamlessly integrating into surrounding human traffic participants\cite{schwarting2018planning}. Fig. \ref{fig:compare} shows a typical scenario where two vehicles encounter at the intersection. Fig. \ref{fig:compare} (a) illustrates the real-world data and Fig. \ref{fig:compare} (b) shows the generated results via a path-planning approach with constrains from road profiles as well as initial/terminal positions and speeds. Although the path-planning algorithm can accomplish the task safely, the generated behavior is still far from similar to human driver's behaviors. One of the main reasons is that human drivers usually make a non-globally optimal decision. Many human-like planning and control methods have been proposed. Yu, \textit{et al.} \cite{yu2018human} developed a human-like lane-changing controller which evaluates the optimal moment and acceleration for changing lanes by estimating the aggressiveness of surrounding vehicles. He, \textit{et al.} \cite{he2018human} formulated a hand-crafted cost function by considering lane incentive and learned the coefficients from on-road lane change data. Besides, some imitation learning and deep learning methods are also used to train controllers and decision-makers of self-driving cars with human demonstrations \cite{muhlig2012interactive, bojarski2016end, wolf2017learning}.


Most of the methods mentioned above need a variety of driving scenarios to train models. However, the multi-vehicle interaction scenarios in the released driving data sets are insufficient \cite{wang2017much}. One of the commonly used methods is to collect data for specific tasks in a driving simulator with different driver participants assuming a certain level of replayability and controllability of simulated driving scenarios \cite{yu2018human}. Another alternative is to collect several data from expert drivers and then build the behavior model from which more data can be generated. Do, \textit{et al.} \cite{do2017human} proposed an active-passive model by analyzing different cases of expert drivers' lane-changing behavior data. The authors discretized the lane change behaviors into five states, i.e., wait, accelerate, decelerate, evade, and lane change. This method is suitable for specific vehicle behaviors but not tractable for modeling driving behaviors in complex scenarios (e.g., multiple vehicles at intersections). In order to solve this problem, deep learning technologies have been implemented. Ding \textit{et al.} \cite{ding2018multi} encoded features of a variety of driving scenarios to latent states then decoded new scenarios by sampling. However, the generation process did not take road geometry constrains and the initial/terminal status of two vehicles into consideration, and also evaluating the generation performance was tricky because of a lack of ground truth. 

Based on the discussion above, it is necessary to develop an efficient approach to generate multi-vehicle scenarios with considering road constrains. Inspired by the empirically proved concept that human driver behavior is composed of countable infinite fundamental building blocks, called {\it traffic primitives}, from which we can generate new scenarios with a stochastic process. To this end, we propose a Gaussian process based approach to generate new multi-vehicle interaction scenarios by integrating traffic primitives with path planning algorithms.
Given a multi-vehicle driving scenario, the proposed method can generate human-like trajectories that fit different road constrains while inheriting the key interaction patterns. Our main contributions are threefold:

\begin{itemize}
    \item Presenting a learning-based framework to extract interpretable traffic primitives from complex multi-vehicle intersection scenarios with less prior knowledge.
    \item Integrating a stochastic process with a reformative path-planning algorithm (i.e., RRT$^{\ast}$-Connect) to generate human-like multi-vehicle interaction trajectories consistent with road constrains.
    \item Evaluating the generation performance by comparing with the desired naturalistic driving scenario and verify the generation results.
\end{itemize}

The remainder of this paper is organized as follows. Section II defines the scenario generation problem and traffic primitive extraction. Section III shows the proposed Gaussian process-based generation approach with traffic primitives. Section IV discusses and analyzes the experimental results. Section V concludes the work and introduces future works.

\section{Traffic Primitive Extraction}
In this section, we will first mathematically define multi-vehicle interaction driving scenarios and then illustrate the methodologies to extract traffic primitives.

\subsection{Multi-Vehicle Interaction Driving Scenario} 
The multi-vehicle driving scenario here is referred as the situation where multiple vehicles are spatially and temporally close to and interact with each other. A complete scenario $\textbf{Y}$ includes states $S$ of each engaged vehicle, which can be described as
\begin{equation}
    \textbf{Y} = \{S^n\}^N_{n=1}
\label{scenario_def}
\end{equation}
where $N$ is the number of engaged vehicles. The state of each vehicle includes positions and speeds ordered by time, which can be written as
\begin{equation}
    S^n = \{s_t^n\}_{t=1}^{T_n} = \{p^n_t, v^n_t\}_{t=1}^{T_n} 
\end{equation}
where $p_t^n \in \mathbb{R}^2$ and $v_t^n \in \mathbb{R}$ represent the position and speed of vehicle $n$ at time $t$, respectively. $T_n$ is the time length of $S^n$. In order to facilitate the analysis and demonstration process, all the vehicles in one scenario are considered as the same time length; that is, $T_n = T$, for $\forall n\in [1,N]$.

\subsection{Traffic Primitives Extraction}

Human driver behavior can be considered as a continuous stochastic process with several potential changepoints (i.e., the points split different states) \cite{nechyba1998stochastic}, thus defining the primitives of driving behaviors. Finding the changepoints of driving scenarios can facilitate the modeling and analysis of driving behaviors, thereby providing basis for the decision-making of autonomous vehicles \cite{joseph2011bayesian, galceran2015multipolicy}. Manually segmenting multi-vehicle driving scenarios with high-dimensional observations is intractable due to limited prior knowledge. Some clustering methods such as Gaussian mixture models (GMM) \cite{gadepally2013framework, havlak2013discrete} can segment driving scenarios spatially, but ignore the temporal dynamics information of time series. Hence, in order to automatically extract traffic primitives in a spatiotemporal space with less prior knowledge, the Bayesian nonparametric learning is introduced by combining a hierarchical Dirichlet process (HDP) \cite{teh2005sharing} with an additional sticky parameter and a hidden Markov model (HMM), denoted as sticky HDP-HMM \cite{fox2011sticky}. The details are shown as follows.

The driving scenario is modeled as a Markov process where the observation of all the vehicles at time $t$ (denoted as $\boldsymbol{s_t}$ = $[s_t^1, s_t^2, ..., s_t^N]^T$) is treated as a sample. Based on the Markov property, we have
\begin{equation}
        \boldsymbol{s_t}|\boldsymbol{s_{t-1}}, \boldsymbol{s_{t-2}}, ..., \boldsymbol{s_{0}} = \boldsymbol{s_t}|\boldsymbol{s_{t-1}}
\end{equation}
indicates that the current observation only depends on the most recent one. Each observation $s_t$ corresponds to a discrete hidden state $z_t$, indicating which kinds of traffic primitives it belongs to. The transition probability from states $z_i$ to $z_j$ is denoted as $\pi_{i,j}$. $\pi_i$ can be considered as a discrete distribution as $\sum_{j=1}^J \pi_{i,j}=1$ where $J \in \mathbb{N}^{+}$ is the number of types of traffic primitives. The observation $s_t$ subject to $z_t$ is drawn from a distribution with parameters $\theta_{z_t}$. 
However, the domain of discrete states may vary over scenarios and might increase with more data samples observed. In order to relax the constrains of HMM in terms of the hidden states and transition probability while ensuring $\sum_{j=1}^{J} \pi_{i,j}= 1, \forall J$, the Dirichlet process (DP) is introduced as $G_0 \sim DP(\gamma, H)$, which can be realized by the stick-breaking process as
\begin{subequations}\label{dp}
    \begin{align}
    G_0 & = \sum_{k=1}^{\infty}\beta_k\delta_{\theta_k},\ \ \theta_k \sim H \\
    \beta_k & = v_k\prod_{\ell=1}^{k-1}(1-v_\ell),\ \ v_i \sim \mathrm{Beta}(1,\gamma)
    \end{align}
\end{subequations}
where $H$ is the base measure, $\delta_{\theta_k}$ is the probability measure concentrated at $\theta_k$. However, the atoms between different $G_0$ are different, even they are sampled from the same base measure $H$. Therefore, the prior distribution of $\pi_i$ is defined via a hierarchical Dirichlet process (HDP)
\begin{subequations}\label{hdp}
    \begin{align}
    \pi_i & \sim DP(\alpha, G_0)\\
    \pi_i & = \sum_{k=1}^{\infty}\pi_{i,k}\delta_{\theta_k}
    \end{align}
\end{subequations}
where $G_0$ is drawn from (\ref{dp}) and $\alpha$ is the concentration parameter.


Based on the algorithms introduced above, we further introduce the sticky HDP-HMM by adding a stick parameter $\kappa \in [0,1]$ to control the transition frequency from one hidden state to other hidden states. A large $\kappa$ would enforce the current state value to be consistent to the next one. Then, the concentration parameter and the base measure in (\ref{hdp}) will be modified as $\alpha+\kappa$ and $\frac{\alpha G_0 + \kappa \delta_{\theta_k}}{\alpha+\kappa}$, correspondingly. The generative model of the sticky HDP-HMM can be formulated by
\begin{subequations}
    \begin{align}
    \theta_k & \sim H \\
    G_0 & \sim DP(\gamma, H) \\
    \pi_i & \sim DP(\alpha+\kappa, \frac{\alpha G_0 + \kappa \delta_{\theta_k}}{\alpha+\kappa}) \\
    z_t|z_{t-1} & \sim \pi_{z_{t-1}} \\
    \boldsymbol{s_t}|z_t & \sim F(\theta_{z_t})
    \end{align}
\end{subequations}
More details can be found in \cite{wang2018understanding, fox2011sticky}. {{The extracted primitives allow us to define the changepoints of sequential states, $s_c$, as first states where the type of primitives changes. Finding out changepoints of complex driving scenarios can help us semantically understand the behaviors behind and benefit the decision-making process by providing state transition information.}}  

\section{Multi-Vehicle Scenario Generation}
In this section, we will propose a method to generate multi-vehicle interaction scenarios with the constraints of road context and vehicle status based on the extracted traffic primitives from a provided scenario, which will carried out through three steps: affine transformation, trajectory planning, and regression. 

\subsection{Affine Transformation}
In real applications, the provided driving scenario can not always fit in any road geometry and additional requirements such as initial/terminal conditions. Thereby, an affine transformation is essential before implementing path planning. In order to make the relative distance between trajectories intact during affine transformation, we only consider rotation, translation, and scaling. More specifically, the template scenario is transformed such that multiple vehicles start and end at target positions $q_{\mathrm{start}}$  and $q_{\mathrm{end}}$, respectively. We use $\overrightarrow{V_q}$ to represent the vector from $q_{\mathrm{start}}$ to $q_{\mathrm{end}}$ and $\overrightarrow{V_p}$ to represent the vector from the starting point $p_{\mathrm{start}}$ to the endpoint $p_{\mathrm{end}}$ of the provided scenarios. Then, the rotation and translation can be formulated as a rigid body transformation matrix
\begin{equation}
    \mathbf{Tr} = 
    \begin{bmatrix}
    \cos \theta_r & -\sin \theta_r & t_x\\
    \sin \theta_r & \cos \theta_r & t_y\\
    0 & 0 & 1
    \end{bmatrix}
\end{equation}
where $t_x$ and $t_y$ represent the translation along $x$-axis and $y$-axis, respectively; $\theta_r$ is the rotation angle, which can be calculated by 
\begin{subequations}
    \begin{align}
    \theta_r & = \cos^{-1}(
    \frac{\overrightarrow{V_p}\cdot \overrightarrow{V_q}}{\|\overrightarrow{V_p}\|_2 \cdot \|\overrightarrow{V_q}\|_2}) \\
    \begin{bmatrix}
        t_x \\ t_y 
    \end{bmatrix} 
    & = \overrightarrow{V_q} - \overrightarrow{V_p}
    \end{align}
\end{subequations}
Thus, the changepoints under the target frame (denoted as $q_c$) is calculated as 

\begin{equation}
    q_c = f_s(\mathbf{Tr}\cdot p_c-q_{\mathrm{start}})+q_{\mathrm{start}}
\end{equation}
with a scaling factor $f_s=\|\overrightarrow{V_q}\|_2/\|\overrightarrow{V_p}\|_2$, where $p_c$ is the trajectory changepoint under the original frame.

\subsection{Trajectory Planning} 
Our task is to generate similar but not identical trajectories, thus a sampling-based path planning method --  RRT$^\ast$-Connect -- is selected, which can efficiently generate plenty of similar, asymptotically optimal trajectories \cite{klemm2015rrt}. The whole algorithm procedure is shown in \textbf{Algorithms} \ref{RRT*-connect}-\ref{rewire} in the \textsc{Appendix}.

The RRT$^\ast$-Connect algorithm combines the benefits of RRT$^\ast$ and RRT-Connect by finding a solution faster than RRT$^\ast$ and more optimal than RRT-Connect. When a collision-free new node $q_{\mathrm{new}}$ is generated based on randomly sampling, in order to minimize the total cost,  a set of near nodes $Q_{\mathrm{near}}$ will be considered as candidate parents instead of directly choosing $q_{\mathrm{nearest}}$. After connecting $q_{\mathrm{new}}$ with $q_{\min}$, we then rewire (\textbf{Algorithm} \ref{rewire}) all $q_{\mathrm{near}} \in Q_{\mathrm{near}}$ if the connection of $q_{\mathrm{near}}$ through $q_{\mathrm{new}}$ would cause lower cost than its previous one. In order to improve the computational efficiency, the range of near nodes $|V|$ is determined by 

\begin{equation}
    |V| = \min(\gamma(\log n)/n)^d, \zeta)
\end{equation}
where $\gamma$ and $\zeta$ are user-defined parameters, $n$ is the number of nodes in both trees, and $d$ is the dimension of searching space. The nearest node in the other tree tries to connect to $q_{\mathrm{new}}$, as shown in \textbf{Algorithm} \ref{connect}. If the extension process encounters obstacles, two trees are swapped as a RRT-Connect algorithm does. The main difference lies in that our algorithm would not be terminated once the path has been found, and inversely, the algorithm will continue exploring and finding more potential paths. The final path can be determined according to \textbf{Algorithm} \ref{RRT*-connect} (lines 7 and 8). The basic idea is that the node $q_{\mathrm{new}}$ would be in two trees when connecting two trees and hereby has two costs, representing the entire costs of the path. Thus, all the $q_{\mathrm{new}}$ in two trees are recorded, and $q_{\mathrm{new}}$ are selected such that the path passing through it obtains the minimum cost.

\subsection{Stochastic Trajectories Generation with GP Regression}

\begin{figure}[t]
\centering
\includegraphics[width=\linewidth]{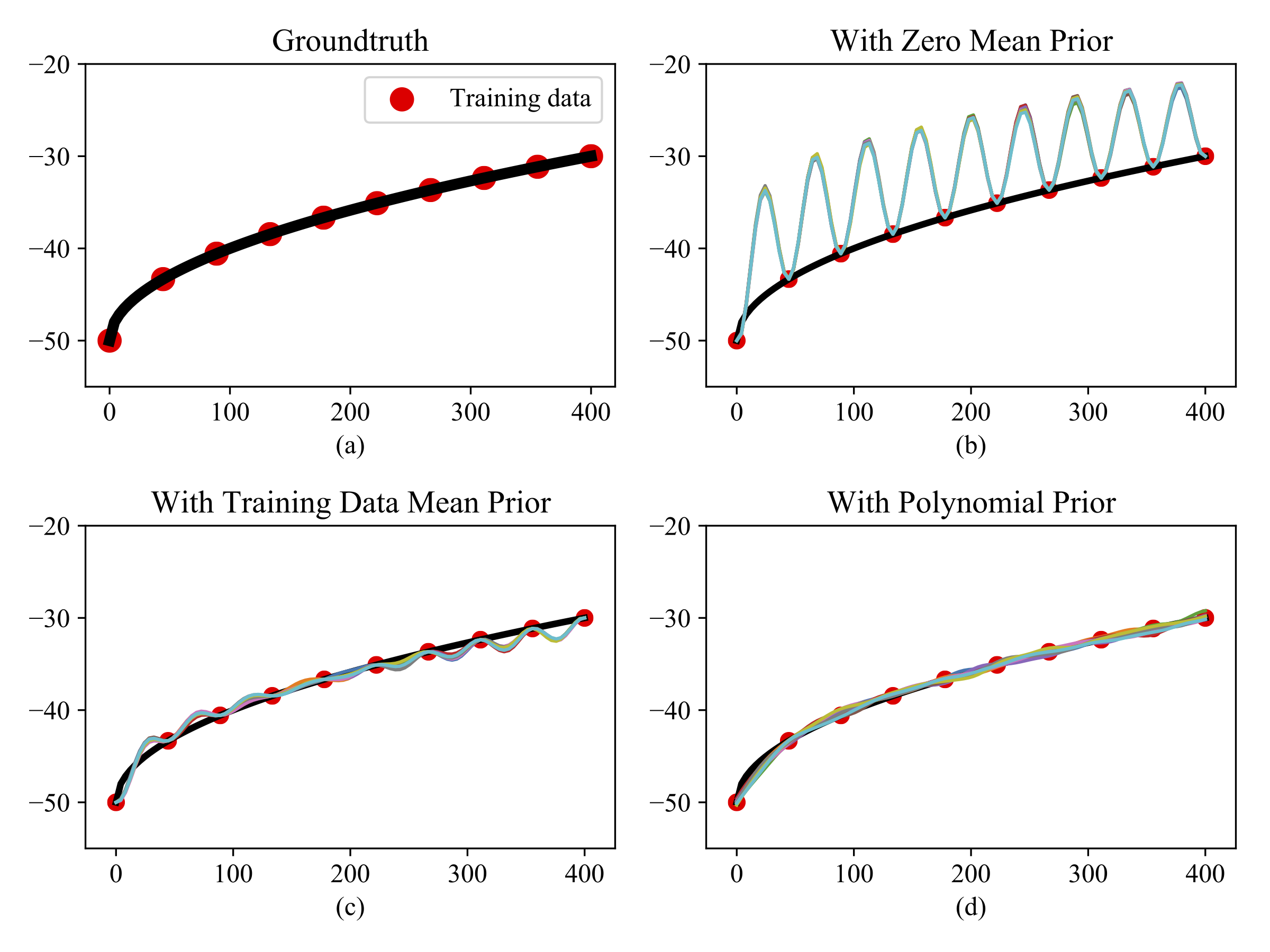}
\caption{The comparison of the GP regression results with different priors. (a) The ground truth (black line) with five training data (red dots). The posteriors (10 regression results sampled for each case) with (b) zero mean prior, (c) training data mean prior and (d) a fitted polynomial function prior.}
\label{fig:gp_cmp}
\end{figure}

After piece-wisely generating trajectories via the changepoints, the regression is essential in order to make the trajectories smooth and then generate stochastic trajectories. One approach is the Gaussian process (GP) since the smooth trajectories can be represented functionally by given a few observations \cite{mukadam2016gaussian}. Besides, a huge bunch of trajectories (with the same means) can be easily sampled. A GP is a collection of random variables where any arbitrary subset is subject to a joint Gaussian distribution. Here, the path generated via RRT$^\ast$-Connect is considered as a 2D curve and hereby described as a Bayesian regression model
\begin{equation}
    y = \underbrace{\phi(x)\omega + p_r(x)}_{f(x)} + \varepsilon
\end{equation}
where $x$ and $y$ represent the ordinate of the path, $\phi(x)$ is the mapping function, $\omega \sim \mathcal{N}(0, \sigma_\omega^2)$ is the weight parameter, and $\varepsilon \sim \mathcal{N}(0, \sigma_n^2)$ is the white noise. Unlike assuming the GP with a zero mean function, we introduce a deterministic function for $x$, denoted as $p_r(x)$, as a nonzero mean prior. The selection of nonzero mean prior follows the fact that the mean of the posterior process of trajectories is far from zero and can vary a lot, thereby laying a strong impact on the generation results. Fig. \ref{fig:gp_cmp} (b)-(d) display a simple GP regression example with different priors and indicates that a suitable prior could make the regression results much more smooth and close to the ground truth. 
Based on the above discussion, the GP can be defined as
\begin{equation}
    f(x) \sim \mathcal{GP}(p_r(x),\phi(x)\sigma_\omega^2\phi(x'))
\end{equation}
The joint distribution of $Y$ and test outputs $f(X_{*})$ can be written as
\begin{equation}\label{joint_dist_gp}
    \begin{bmatrix}
    Y \\
    f(X_{*})
    \end{bmatrix}
    \sim
    \mathcal{N}\left(
    \begin{bmatrix}
    p_r(X) \\
    p_r(X_{*})
    \end{bmatrix},
    \begin{bmatrix}
    \bar{K}_{\omega} & K(X, X_{*}) \\
    K(X_{*}, X) & K(X_{*}, X_{*})
    \end{bmatrix}\right)
\end{equation}
with $\bar{K}_{\omega} = K(X,X)+ \sigma^2_\omega I$. $X$, $X_*$, and $Y$ represent the corresponding training inputs, test inputs, and training outputs. $K(X, X')$ is the squared exponential covariance function

\begin{equation}
    K(X, X') = \sigma_f^2\exp\left(-\frac{1}{2l^2}|X-X'|^2\right)
\end{equation}
where $\sigma_f$ and $l$ are user-defined hyperparameters to adjust the variance and smoothness. Therefore, the distribution of $f(X_*|X, X_*, Y)$ can be derived from (\ref{joint_dist_gp}) and thereby generate similar scenarios via sampling techniques.

\begin{subequations}\label{gp_posterior_distrib}
\begin{align}
f(X_*) & \sim \mathcal{N}(\mathbb{E}(f(X_*)),  \mathrm{Cov}(f(X_*))) \\
\mathbb{E}(f(X_*))& =p_r(X_*)+K(X_*,X) \bar{K}_{n}^{-1}(Y-p_r(X))    \\
\mathrm{Cov}(f(X_*)) & =K(X_*, X_*)−K(X_∗,X) \bar{K}_{n}^{-1}K(X,X_*)
\end{align}
\end{subequations}
with $\bar{K}_{n} = K(X,X) + \sigma_n^2I$.

\section{Experiment and Data Collection}

\subsection{Data Collection}
The naturalistic driving scenarios we used were collected by the University of Michigan Safety Pilot Model Development (SPMD) program. The database covers more than 10 thousand vehicle-to-vehicle (V2V) driving scenarios from around 3,500 equipped vehicles for more than three years. In order to collect interactive behaviors between vehicles, the dedicated short-range communication (DSRC) technology has been implemented and can be activated when two vehicles are spatial close to each other (less than 100 m). The position (longitude and latitude) and speed information of two vehicles are collected via the onboard GPS and by-wire speed sensor, respectively.

\begin{figure}[t]
\centering
\includegraphics[width=\linewidth]{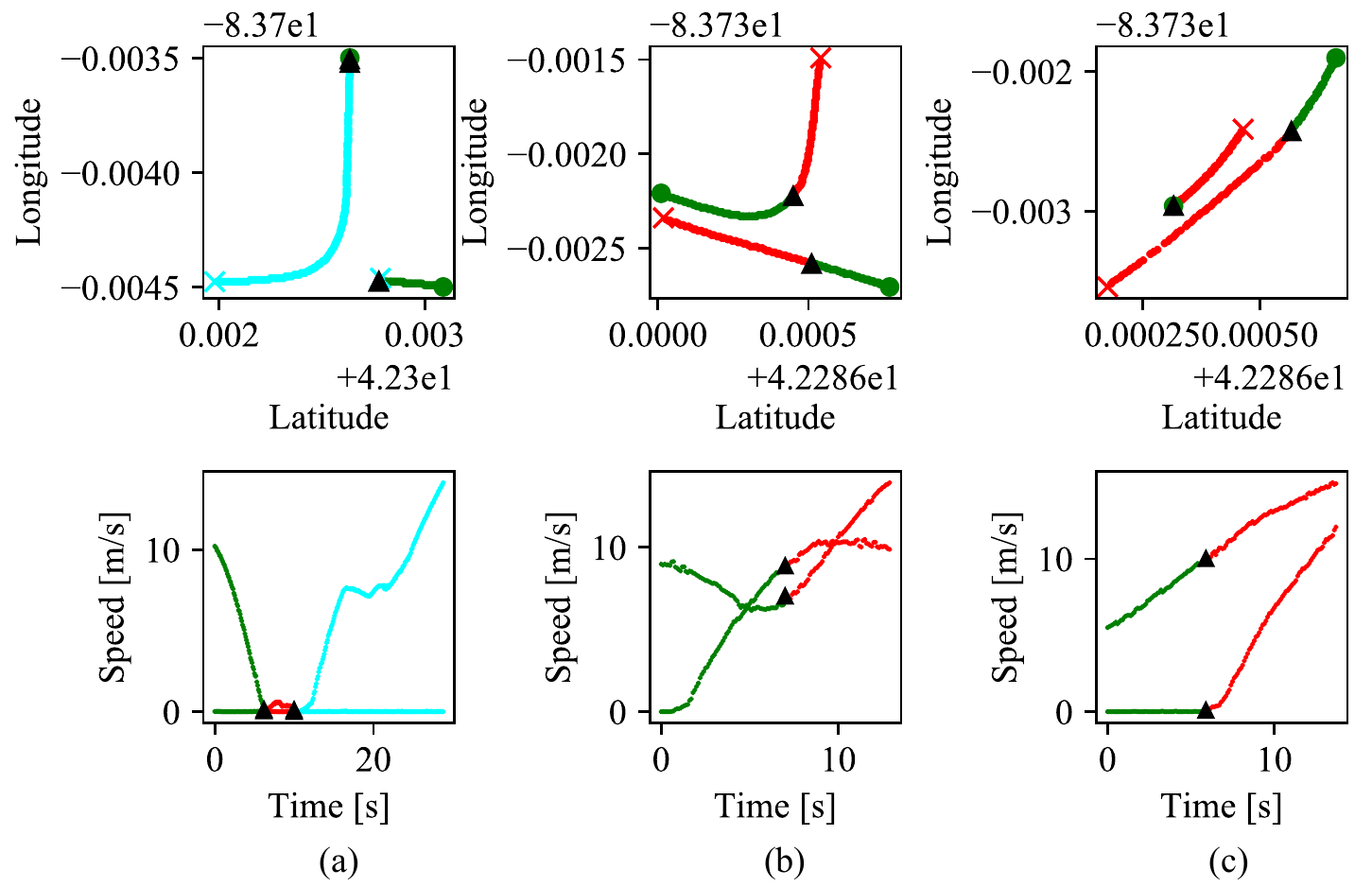}
\caption{Traffic primitive extraction results of three typical driving scenarios. Dot, cross, and upper-triangle are the starting point, endpoint, and changepoints between traffic primitives, respectively.}
\label{fig:hdp_hmm_extrac_rst}
\end{figure}

\begin{figure*}[th!]
\centering
\includegraphics[width=0.93\linewidth]{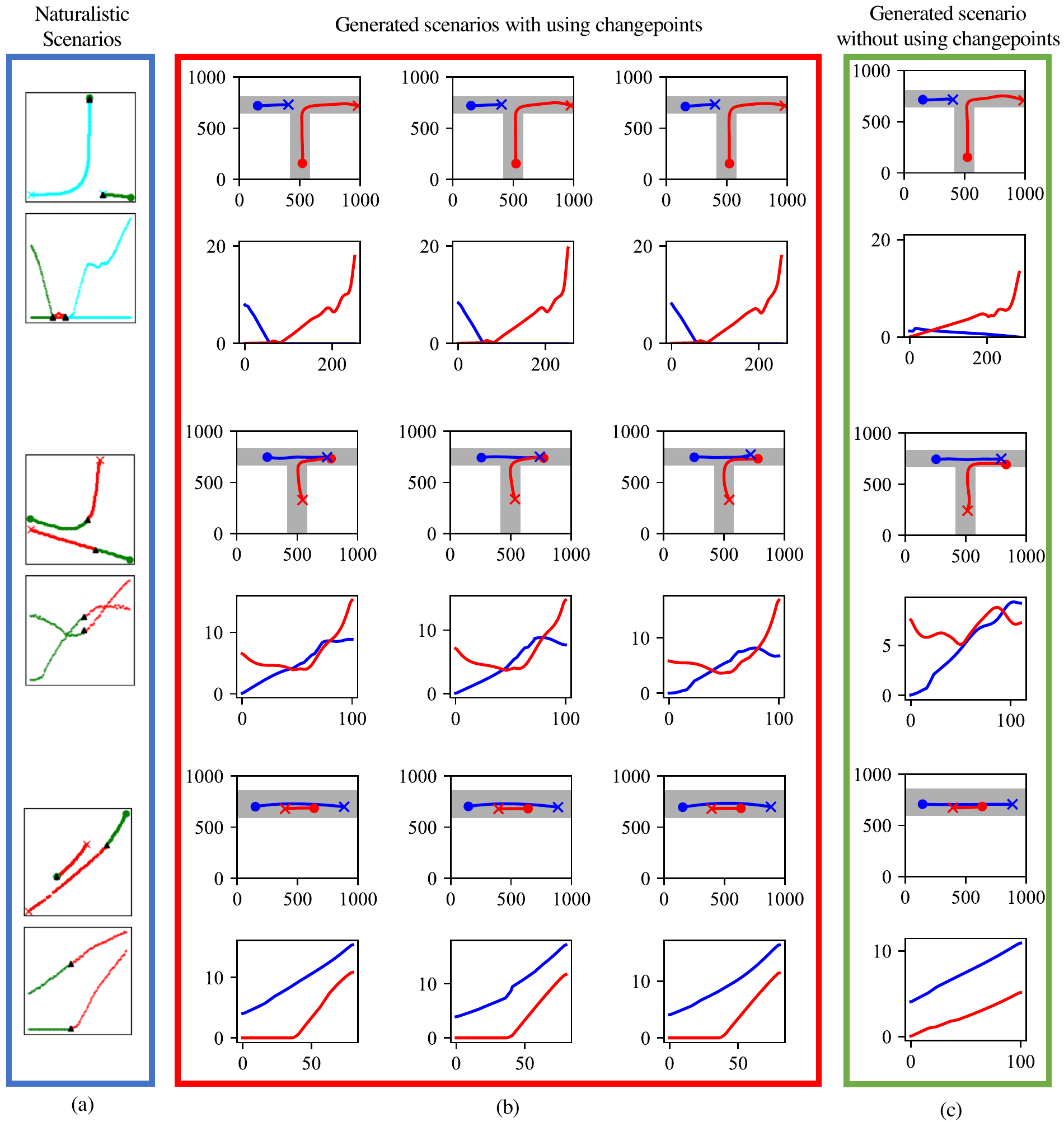}
\caption{Generation results of three typical driving scenarios under different road constrains, where trajectory profiles (upper plots) are plotted in the 1000$\times$1000 grid map and their speed profiles (lower plots) are calculated by differentiation, respectively.}
\label{fig:generation_results}
\end{figure*}

\subsection{Experiment Setting}
The iteration step during training the sticky HDP-HMM is set as 200. The clusters containing less than 2 data points are moved, i.e., the intervals between changepoints should be larger than 0.2 s. In order to show the generation performance under different situations, the straight lane and T-intersection in a 1000$\times$1000 grid map are used. The constraints include road boundary, initial/terminal conditions of position and speed. The step size of the RRT$^\ast$-Connect is 5. The training and testing data of the GP regression are chosen by considering the motion between two changepoints are constant-acceleration movement. We set the hyperparamters $\sigma_\omega = 10$, $\sigma_f = 10$, and $l = 100$. A third order polynomial function was employed on the training data and then considered as the prior function $p_r(\cdot)$. 

\section{Results and Analysis}
\subsection{Traffic Primitive Extraction}

Fig. \ref{fig:hdp_hmm_extrac_rst} displays the traffic primitive extraction results of three typical multi-vehicle interaction driving scenarios, wherein (a) and (b) occur at intersections while (c) occur on the urban straight road. The colors in one scenario distinguish the extracted traffic primitives. It can be seen that the sticky HDP-HMM can automatically find interpretable changepoints of multi-dimensional sequential data over trajectories and speed profiles. Based on the extraction results and analysis, two main conclusions can be drawn as follows.

First, the extracted traffic primitives are interpretable. Fig. \ref{fig:hdp_hmm_extrac_rst} (a) shows a complex scenario that two vehicles negotiate at an intersection. The entire scenarios can be semantically interpreted via the three traffic primitives: 1) First, one vehicle keeps stationary while the other one decelerates for about 7 seconds until being stationary; 2) Then, these two vehicles keep stationary at the same time for about 3 seconds; 3) Finally, one vehicle starts to accelerate to turn right while the other one still keeps stationary. For Fig. \ref{fig:hdp_hmm_extrac_rst} (b), the changepoints (marked as black triangles) between two traffic primitives interpret the interaction behaviors as 1) First, these two vehicles drive in the opposite direction with one vehicle accelerating and the other slowing down and 2) then one vehicle maintains its direction and continues accelerating while the other one turns left and speeds up. Fig. Fig. \ref{fig:hdp_hmm_extrac_rst} (c) can be explained in the same semantic way with the extracted traffic primitives.

Second, the speed is essential for traffic primitive extraction. Fig. \ref{fig:hdp_hmm_extrac_rst} (c) displays a scenario in which two vehicles drive in opposite direction. Without speed profiles, it is difficult to find the changepoints solely using the trajectories since the entire scenario would be considered as a single behavior. Considering speed information can enable us to segment one vehicle's behavior into two clear stages: first keeps still, and then accelerates. 

Besides, the bottom plots (a)-(c) in Fig. \ref{fig:hdp_hmm_extrac_rst} indicate that the changepoints of traffic primitives are usually located at the place where the trend of vehicle speed changes; that is, the trend of speed within each single traffic primitive is identical. Therefore, the acceleration within each traffic primitive can be generally considered as a constant (i.e., minus for deceleration, positive for acceleration, and zero for constant speed). This property allows us to draw data samples from the GP regression based on their acceleration. 

\subsection{Generation Results and Evaluation}
Based on the extracted traffic primitives and their changepoints, we implement our proposed scenario generation method that integrated RRT$^\ast$-Connect and GP regression and then obtain the results as shown in Fig. \ref{fig:generation_results}. Top two scenarios display the generation results of two common driving scenarios occurring at intersections, and the bottom one describes two vehicles encounter and cross with each other on a straight road. In order to analyze the effects of changepoints, Fig. \ref{fig:generation_results} (b) and (c) display the generation results with and without using changepoints, respectively. Comparison results demonstrate that the utlization of changepoints of traffic primitives can indeed capture the key underlying interaction patterns. For instance, the  generated results of the blue vehicle (blue line in the top plot of Fig. 4 (b)) using the changepoints can capture the driver behavior of slowing down to approach to the intersection and then keeping stationary; while the ones without using changepoints (red line in the top plot of Fig. 4 (b)) can not realize the stop behavior during interaction. In each scenario, the upper plots display the generated trajectories in a grid map with size of 1000$\times$1000, from which the speed profiles (lower plots) can be derived via the Euler differentiation method. More specifically, each speed point is calculated by 
\begin{equation}
    v_t = \frac{p_{t+d_t}-p_t}{d_t}
\end{equation} 
  
This paper mainly emphasizes on the scenario generation and performance analysis. Without losing generality, we set $d_t=1$. In the future implementation, the values of trajectory and speed data can be scaled and transformed such that they can be consistent with user's unit system, e.g. m/s and mph.  

Based on the generation performance, several conclusions can be drawn as follows. 
\begin{itemize}
    \item The generation results take into consideration of various additional constraints such as road boundaries and the initial/terminal states of vehicles' position and speed.
    \item The generation results maintain properties of original provided sequential data, and the states of the trajectory and speed at any time can be queried from the GP regression.
    \item The changepoints of traffic primitives endow the generated trajectory and speed profiles with the capability of inheriting the key underlying interaction features in original driving scenarios. Without using changepoints, the generated scenarios have high discrepancy with the naturalistic one even though the road constrains were considered.
 
\end{itemize}

\begin{figure*}[t]
\centering
\includegraphics[width=\linewidth]{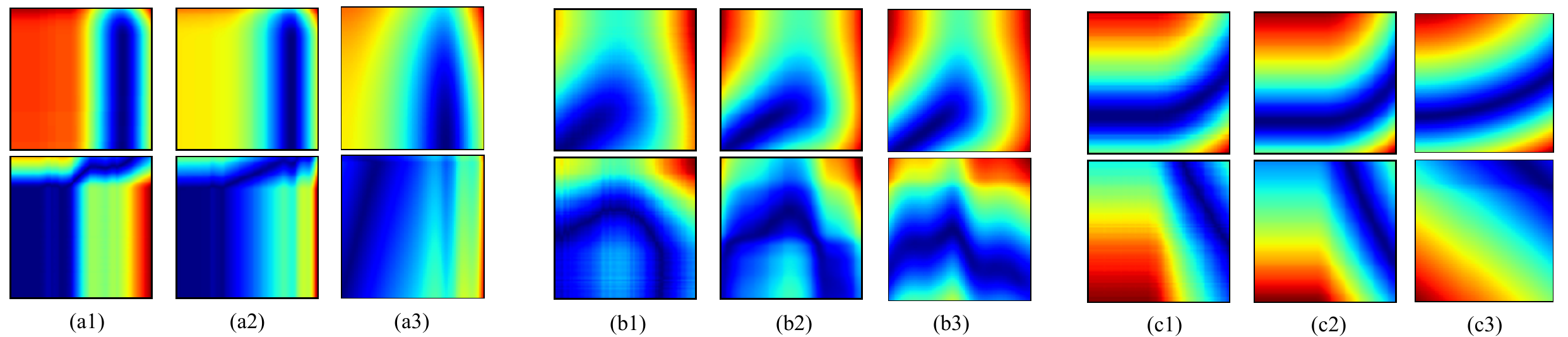}
\caption{The normalized DTW feature of trajectory (top) and speed (bottom) of the generated results with using changepoints ((a2), (b2), (c2)), generated results without using changepoints ((a3),(b3), (c3)) and the provided multi-vehicle interaction scenarios ((a1), (b1), (c1)) in Fig. \ref{fig:hdp_hmm_extrac_rst} in the same order.}
\label{fig:heatmap}
\end{figure*}

In order to compare the generated scenarios with the provided scenarios intuitively, we introduce a measure (i.e., Dynamic Time Warping (DTW)) \cite{muller2007dynamic} to capture the spatiotemporal dynamic interaction between two vehicles via calculating distances or similarity between  temporal sequences in terms of vehicle position and speed. The distance calculation of trajectory and speed depends on their dimension. Here, we use the Euclidean distance for trajectory and the Manhattan distance for speed. Finally, for each provided driving scenario, 50 scenarios are generated. Fig. \ref{fig:heatmap} displays the averages of the normalized DTW feature matrices (top for trajectories and bottom for speed) of the 50 generated multi-vehicle interaction scenarios for each provided driving scenario. Plots (a1)-(c1) are the DTW features of provided scenarios. Plots (a2)-(c2) and (a3)-(c3) are the averaged DTW features of the 50 generated results with and without using changepoints, respectively. Red represents high discrepancy (large distance) while dark blue represents the small difference (small distance). 

From Fig. \ref{fig:heatmap}, we know that the interaction patterns of generated scenarios, represented by the DTW features, can represent the interaction similarity between scenarios. For example, the interaction of the provided and generated two-vehicle driving scenarios (reflected by the DTW features in (a1) and (a2)) are close to each other. However, the color at left parts is slightly different, which indicates that the generated scenario has a closer distance than the provided one because of differences in road constraints. However, the blue pattern in (a3) trajectory profile is much different with (a1), which indicates that the generated scenario without using changepoints have high discrepancy with the template scenario, i.e., can not capture the key underlying interaction information of the provided driving scenarios. Comparison of (c3) with (c1) indicates that the generated scenarios without using changepoints can capture the interaction patterns of trajectory (top plots of (c1) and c(3)), but fail to represent the interaction information of speed (bottom plots of c(1) and c(3)).

In addition, our developed scenario generation method can obtain smoother results. For example, the speed heat map of (c2) is smoother than (c1), indicating that the generated scenarios are more continuous than the provided data because the provided data is collected from sensors with noise.

The above discussion demonstrates that extracting the changepoints of traffic primitives is the key to generating scenarios that inherit the interaction patterns of the given scenarios. This is because the changepoints represents the changes of primitives and the constant acceleration property can be held within each primitives. Therefore, without using the changepoints, the data drawn from GP regression cannot be considered as time sequence and thereby, the generated results will display divergent behaviors with the original one and can not reflect proper interactions of vehicles.

\section{Conclusion}
This paper presents powerful algorithms to generate human-like multi-vehicle interaction scenarios, which can fit different road conditions while inheriting the key interaction characteristics of the loggged naturalistic multi-vehicle interaction behaviors. This proposed approach is based on the concept that human driving behavior is generated from some semantic traffic primitives that can be learned via the Bayesian nonparametric statistics without any prior knowledge required. After fitting the traffic primitives into the desired roads, a bunch of new scenarios that have the similar dynamic interaction pattern with the provided one can be generated by combining a sample-based path planning algorithm (i.e., RRT$^\ast$-Connect) with the Gaussian process regression. Then, the distance-based feature measurement (i.e., DTW) was introduced to evaluate the generation performance. In this work, we successfully generated different kinds of typical driving scenarios occurring at different roads with initial and terminal constraints of vehicle states. The generated scenarios could be used to simulate human drivers behaviors under different conditions in our future work. 

\section*{Appendix}
\begin{algorithm}
\caption{RRT$^\ast$-Connect}
\label{RRT*-connect}
$\mathcal{T}_a \leftarrow \{q_{\mathrm{start}}\}$, $\mathcal{T}_b \leftarrow \{q_{\mathrm{end}}\}$\;
    \For{$k=1$ \KwTo $K$}{
        $q_{\mathrm{rand}} \leftarrow$ Random\_Sample()\;
        \If{$\texttt{Extend} (\mathcal{T}_a, q_{\mathrm{rand}}) \neq \texttt{Trapped}$}{ $\texttt{Connect}(\mathcal{T}_b, q_{\mathrm{new}}$)\;
        }
        $\texttt{Swap}(
        \mathcal{T}_a, \mathcal{T}_b$)\;
    }
$q_{\mathrm{opt}} \leftarrow \operatorname*{\arg \min}\limits_{q\in \mathcal{T}_a \cap \mathcal{T}_b} \texttt{Cost}(q\in{\mathcal{T}_a}) + \texttt{Cost}(q\in{\mathcal{T}_b})$ \;
\Return Path($\mathcal{T}_a, \mathcal{T}_b, q_{\mathrm{opt}})$ \;
\end{algorithm}
%
\begin{algorithm}
\caption{\texttt{Extend}($\mathcal{T}, q$)}
\label{extend}
$q_{\mathrm{nearest}} \leftarrow \texttt{Nearest\_Neighbor}(\mathcal{T}, q$) \;
$q_{\mathrm{new}} \leftarrow \texttt{Steer}(q, q_{\mathrm{nearest}}$)\;
\If{$\texttt{ObstacaleFree}(q_{\mathrm{new}})$}{
    $Q_{\mathrm{near}} \leftarrow \texttt{Near} (\mathcal{T}, q_{\mathrm{new}}, |V|)$ \;
    $q_{min} \leftarrow \texttt{ChooseParent}(Q_{\mathrm{near}}, q_{\mathrm{nearest}}, q_{\mathrm{new}})$ \;
    $\mathcal{T} \leftarrow \texttt{AddNode}(q_{\min}, q_{\mathrm{new}}, \mathcal{T})$ \;
    $\mathcal{T} \leftarrow \texttt{ReWire}(\mathcal{T}, Q_{\mathrm{near}}, q_{\min}, q_{\mathrm{new}})$\;
    \If{$q_{\mathrm{new}} = q$}{
    \Return \texttt{Reached} \;}
    \Else{
    \Return \texttt{Advanced} \;
    }
}
\Return \texttt{Trapped} \;
\end{algorithm}
\begin{algorithm}[h!]
\caption{\texttt{Connect}($\mathcal{T}, q$)}
\label{connect}
 \Repeat{$\textup{S} \neq \textup{Advanced}$}{
      $\textup{S} \leftarrow \texttt{Extend}(\mathcal{T}, q)$\;}
\Return $\textup{S}$ \;
\end{algorithm}
\vspace{-0cm}
\begin{algorithm}[h!]
\caption{\texttt{ChooseParent}($Q_{\mathrm{near}}, q_{\mathrm{nearest}}, q_{\mathrm{new}}$)}
\label{chooseparent}
$q_{\min} \leftarrow q_{\mathrm{nearest}}$ \;
$c_{\min} \leftarrow \texttt{Cost}(q_{\mathrm{nearest}}) + \texttt{Dist}(q_{\mathrm{new}}, q_{\mathrm{nearest}})$\;
\ForEach{$q_{\mathrm{near}} \in Q_{\mathrm{near}} \setminus q_{\mathrm{nearest}}$}{
    \If{$\texttt{ObstacaleFree}(q_{\mathrm{near}}, q_{\mathrm{new}}) \And{\texttt{Cost}(q_{\mathrm{near}}) + \texttt{Dist}(q_{\mathrm{new}}, q_{\mathrm{near}})} \leq c_{\min}$}{
        $q_{\min} \leftarrow q_{\mathrm{near}}$ \;
        $c_{\min} \leftarrow \texttt{Cost}(q_{\mathrm{near}}) + \texttt{Dist}(q_{\mathrm{new}}, q_{\mathrm{near}})$\;
    }
}
\Return $q_{\min}$ \;
\end{algorithm}
\vspace{-0cm}
\begin{algorithm}[h!]
\caption{\texttt{ReWire}($\mathcal{T}, Q_{\mathrm{near}}, q_{\min}, q_{\mathrm{new}}$)}
\label{rewire}
\ForEach{$q_{\mathrm{near}} \in Q_{\mathrm{near}} \setminus q_{\min}$}{
    \If{$\texttt{ObstacleFree}(q_{\mathrm{near}}, q_{\mathrm{new}}) \And{\texttt{Cost}(q_{\mathrm{near}}) > \texttt{Cost}(q_{\mathrm{new}}) + \texttt{Dist}(q_{\mathrm{new}}, q_{\mathrm{near}})}$}{
        $\texttt{Cost}(q_{\mathrm{near}}) \leftarrow \texttt{Cost}(q_{\mathrm{new}}) + \texttt{Dist}(q_{\mathrm{new}}, q_{\mathrm{near}})$ \;
        \texttt{ReConnect}($q_{\mathrm{near}}$, $q_{\mathrm{new}}$. $\mathcal{T}$) \;
    }
}
\end{algorithm}

\section*{Acknowledgment}
\label{sec: ack}

Toyota Research Institute (TRI) provided funds to assist the authors with their research but this article solely reflects the opinions and conclusions of its authors and not TRI or any other Toyota entity. 

\bibliographystyle{IEEEtran}
\bibliography{references}

\end{document}